\documentclass{article}
\usepackage{spconf,amsmath,graphicx,hyperref,booktabs}
\usepackage{multirow}
\title{LibriSpeech-Real-QA: A Benchmark for Streaming Question Answering}
\name{\shortstack{
Zewen Zhan$^{1*}$,
Hui Lu$^{2*}$,
Xueyuan Chen$^{2}$,
Huimeng Wang$^{2}$,
Yikai Huang$^{1}$ \\
Xixin Wu$^{2}$,
Shiyin Kang$^{3}$, 
Bin Xia$^{3}$,
Zhiyong Wu$^{1}$
}}
\address{
$^{1}$Tsinghua University, China\\
$^{2}$The Chinese University of Hong Kong, China\\
$^{3}$SenseTime Research, China\\
\texttt{zhanzw25@mails.tsinghua.edu.cn, luhui@se.cuhk.edu.hk}
}
\begin{document}
\ninept
\maketitle
\let\thefootnote\relax\footnotetext{$*$These authors contributed equally to this work.}
\footnotetext{This work was done at SenseTime Research.}
\footnotetext{A live demo is available at \url{https://zxzxz-1.github.io/librispeech-real-qa-demo/}.}

\begin{abstract}
Most speech large language models (SLMs) still process speech in an offline manner --- they only start responding after they receive the complete speech segment.
However, many real-world applications require streaming speech interaction, for which the audio continuously flows in and the machine is required to answer questions based on the so far received audio chunks. For example, a user may ask the system to continuously count how many times a person’s name has been mentioned while they are speaking.
We observe a lack of benchmarks and evaluation protocols for real-time speech interaction. Motivated by this, we propose LibriSpeech-Real-QA, a dataset and a benchmark for real-time speech interaction. We construct different types of real-time question-answer pairs from LibriSpeech to evaluate whether speech LLMs can identify target words in streaming audio. We evaluate several popular real-time commercial models and find that they still tend to respond only after the speech turn has finished. We demonstrate that we can enable a pretrained SLM to perform this task, where we explore several modeling designs and establish a solid baseline for future research.
\end{abstract}

\section{Introduction}
\label{sec:intro}
In a human spoken conversation, interlocutors listen to and understand the speech from each other before it is finished. We argue that machines require the same streaming speech interaction capability to better interact with humans in speech.
Speech large language models (SLMs) have demonstrated impressive capabilities in speech understanding, generation and spoken dialogue ~\cite{DBLP:journals/corr/abs-2407-10759,DBLP:conf/icml/KongGBPVC24,DBLP:journals/corr/abs-2504-18425,DBLP:journals/corr/abs-2412-02612,DBLP:conf/iclr/FangGZMZ025}. However, these models predominantly operate in a strictly turn-based setting, where the complete audio signal needs to be fully available before processing begins.

Full-duplex spoken interaction \cite{DBLP:journals/corr/abs-2410-00037,veluri-etal-2024-beyond,DBLP:journals/corr/abs-2505-17060,DBLP:conf/interspeech/HuHCCGZCLBG25,zhang-etal-2025-omniflatten,DBLP:journals/corr/abs-2605-10199} is promising to facilitate streaming speech interaction, as it requires models to automatically decide their dialogue state in real-time, i.e., to continue listening, continue speaking, convert from listen to speak or vice versa.
However, the typical evaluation scenarios are turn-taking, user interruption and backchannels \cite{DBLP:conf/asru/LinLLWALL25,lin2025fdb_v15}, and there is a lack of explicit evaluation regarding fine-grained streaming speech interaction capabilities, such as quick assistant responses during streaming speech.

More recently, speech interactions have evolved beyond the full-duplex spoken dialogue towards more general real-time speech and audio interaction. Audio Interaction model \cite{DBLP:journals/corr/abs-2606-05121} proposes Proactive-Sound-Bench to evaluate a series of audio-event-triggered proactive response tasks. Thinking Machines Lab reports an interesting internal evaluation task for their Interaction Models \cite{thinkingmachines2026interactionmodels} to identify certain words in an audio stream in real-time. We believe that such curated streaming speech interaction tasks can serve as an important indicator for the SLM in terms of their streaming speech processing and real-time interaction capabilities, but we have not seen open benchmarks regarding these capabilities.

Traditional streaming speech processing tasks, such as streaming speech recognition~\cite{DBLP:conf/icassp/JiaKZL0WSMK25} and translation~\cite{DBLP:conf/nips/LeHLCQ25}, can also serve as proxy tasks for streaming speech interaction. However, they emphasize more on the passive processing of the speech signal, leaving proactive interaction capabilities of the SLMs untested.

To address these gaps, we introduce \textbf{LibriSpeech-Real-QA}, a benchmark for streaming speech interaction and question answering constructed from LibriSpeech \cite{DBLP:conf/icassp/PanayotovCPK15} corpus. The benchmark covers four types of tasks that follow the same format: the model receives a speech stream together with a prompt that asks it to emit an answer when the queried event is heard. The types cross a lexical or semantic target with detection or counting (Section~\ref{sec:tasks}). They allow us to evaluate real-time speech understanding, instruction following, and proactive response generation. We propose a corresponding metric calculation protocol to evaluate streaming SLMs' predictions with emission-level recall, precision, F1-score, and onset delay. Finally, we present a finetuning approach for adapting a pretrained SLM---Qwen2.5-Omni \cite{DBLP:journals/corr/abs-2503-20215}---to our proposed streaming speech interaction tasks, establishing strong baselines for future research.
\section{Benchmark Design}
\label{sec:benchmark}

\subsection{Task Definitions}
\label{sec:tasks}

We define four streaming question answering tasks. In each task, a model receives an audio stream along with a natural language prompt and must output natural language tokens at appropriate times.

\begin{itemize}
\setlength{\topsep}{2pt}
\setlength{\partopsep}{0pt}
\setlength{\itemsep}{0pt}
\setlength{\parsep}{0pt}
\setlength{\parskip}{0pt}

\item \textbf{Lexical Detection.} Given a target word $w$, the model must output $w$ immediately each time it hears $w$ in the audio stream.
\item \textbf{Lexical Counting.} Given a target word $w$, the model maintains a running count and outputs the updated count immediately after each occurrence of $w$.
\item \textbf{Semantic Detection.} Given a semantic class $c$ (e.g., ``animal,'' ``city''), the model must output each word as soon as it hears a word belonging to class $c$.
\item \textbf{Semantic Counting.} Given a semantic class $c$, the model maintains a running count and outputs the updated count whenever a word belonging to $c$ is heard.

\end{itemize}
\subsection{Data Construction}
\label{sec:data}
We construct a dataset for the proposed tasks from the LibriSpeech corpus~\cite{DBLP:conf/icassp/PanayotovCPK15}. As shown in Table~\ref{tab:splits}, it follows the LibriSpeech partition: a training set constructed from \texttt{train-clean-100}, \texttt{train-clean-360}, and \texttt{train-other-500}; a development set from \texttt{dev-clean} and \texttt{dev-other}; and a benchmark from \texttt{test-clean} and \texttt{test-other}. Each utterance is annotated with three domains: the system prompt, the list of target words, and their timestamps. The annotations are produced by a pipeline of four stages as described below.

\textit{Stage 1: Forced Alignment.} We obtain word-level timestamps using a force-aligner such as Montreal Forced Aligner (MFA) alignments~\cite{mfa}. Each word is assigned a start time and end time.

\textit{Stage 2: Semantic Classification.} We classify words into one of 20 semantic categories (including \textit{person\_name}, \textit{animal}, \textit{city}, \textit{profession}, \textit{organization}, \textit{food}, \textit{plant}, \textit{color}, \textit{body\_part}, \textit{cardinal}, \textit{ordinal}, \textit{kinship}, \textit{date}, \textit{clock\_time}, \textit{country}, \textit{money}, \textit{disease}, \textit{weather}, \textit{state\_or\_province} and \textit{Null}) using an LLM with a structured extraction prompt. Classified semantic labels are stored alongside timestamps in the metadata.

\textit{Stage 3: Sample Generation.} For each utterance, we select target words and semantic classes that appear in the metadata, and we fill a prompt template with the chosen target. The prompt is the instruction for one of the four tasks in Section~\ref{sec:tasks}. A lexical-detection example is: ``The moment you hear the word `warriors' in the stream, say it out loud.'' When a target occurs more than once, we keep the sample only if consecutive occurrences are separated by more than one second, leaving the downstream model ample time to emit an answer before the next event. Lexical and semantic counting samples are built only when the target occurs multiple times.

\textit{Stage 4: Negative Sample Generation.} The samples from Stage 3 are positive: the utterance always contains the word or semantic class named in the prompt. We also build negative samples, in which that target is absent, so the correct behavior for the model is to remain silent. Without exposing to negative samples, a model can treat every utterance as one that requires an answer while including them in training reduces that bias.

\begin{table}[htb]
\centering
\caption{Number of samples by split and task.
Lex.\ det.: lexical detection; Lex.\ cnt.: lexical counting;
Sem.\ det.: semantic detection; Sem.\ cnt.: semantic counting.}
\label{tab:splits}
\resizebox{\columnwidth}{!}{%
\begin{tabular}{llrrrr}
\toprule
Set & Split & Lex.\ det. & Lex.\ cnt. & Sem.\ det. & Sem.\ cnt. \\
\midrule
\multirow{3}{*}{Train}
  & train-other-500 & 138{,}800 & 38{,}282 & 199{,}111 & 38{,}889 \\
  & train-clean-360 & 92{,}593  & 28{,}804 & 137{,}485 & 27{,}584 \\
  & train-clean-100 & 26{,}529  & 7{,}639  & 40{,}260  & 8{,}014  \\
\midrule
\multirow{2}{*}{Dev}
  & dev-clean       & 2{,}511   & 243    & 2{,}437   & 309    \\
  & dev-other       & 2{,}623   & 239    & 2{,}300   & 260    \\
\midrule
\multirow{2}{*}{Benchmark}
  & test-clean      & 2{,}399   & 237    & 2{,}392   & 277    \\
  & test-other      & 2{,}658   & 270    & 2{,}636   & 289    \\
\bottomrule
\end{tabular}%
}
\end{table}

\subsection{Evaluation Protocol}
\label{sec:eval}
The benchmark evaluates models that take streaming speech input chunk by chunk, with chunk duration $C$, and emit a response as soon as the queried target is heard. Both the time and the content of the response are scored in the evaluation.

Each occurrence of the target word is a reference event. For detection the target content is the word text, and for counting it is the updated count at that word. The event occupies the aligned interval $[t_{\mathrm{start}}, t_{\mathrm{end}}]$ in the streaming speech. 
For each event, we define its acceptance window as $[t_{\mathrm{start}},\, t_{\mathrm{end}}+g]$. It opens when the word starts and remains open for a grace period $g$ after the word ends in the stream speech.
A model emission is considered as a hit only when its onset ($\hat{t}$) is within the acceptance window of the event and its content agrees with the event and the prompt.

We use $N_{\mathrm{ref}}$ to denote the number of reference events and $\mathrm{TP}$ for the number of hits. A \emph{miss} is a reference event that receives no hit, so the number of misses is $N_{\mathrm{ref}}-\mathrm{TP}$. A \emph{false alarm} ($\mathrm{FA}$) is a model emission that matches no event.

Before matching, we lowercase text strings and remove non-alphanumeric characters, and we treat number words as equivalent to digits, so that \texttt{two} matches \texttt{2}. We report \emph{fuzzy} matching, under which the reference text only needs to occur inside the prediction. \emph{Exact} matching requires the two normalized strings to be identical.

We calculate recall, precision, and F1 scores over all events in a split, with $\mathrm{Recall}=\mathrm{TP}/N_{\mathrm{ref}}$ and $\mathrm{Precision}=\mathrm{TP}/(\mathrm{TP}+\mathrm{FA})$. We also calculate the mean onset delay $\overline{d}$ which averages $\hat{t}-t_{\mathrm{start}}$ over hits. Counting uses the same matching rule, with each count update as its own event, so both the count and the onset must be correct.

\section{Streaming Model Baseline}
\label{sec:baseline}

We adapt Qwen2.5-Omni-7B~\cite{DBLP:journals/corr/abs-2503-20215} to perform the proposed tasks by a sequence of causal decisions at chunk boundaries: after each fixed-duration chunk the model either stays silent (by emitting a special silence token) or emits the target content once the queried event becomes observable.

\subsection{Modeling Formulation}
\label{sec:streaming_formulation}

We divide an audio stream into $K$ consecutive chunks $x_1,\ldots,x_K$ each lasting $C$ seconds, represented in the native multi-turn format of Qwen2.5-Omni as $[\mathrm{system}],\{[\mathrm{user}(x_k),\mathrm{assistant}(y_k)]\}_{k=1}^{K}$. Each user turn contains only the newly arrived chunk, so predicting $y_k$ accesses only the causal prefix $x_1,\ldots,x_k$.

\begin{figure*}[!t]
\centering
    \includegraphics[
        width=0.75\textwidth,
        trim=0 210 0 100,
        clip
    ]{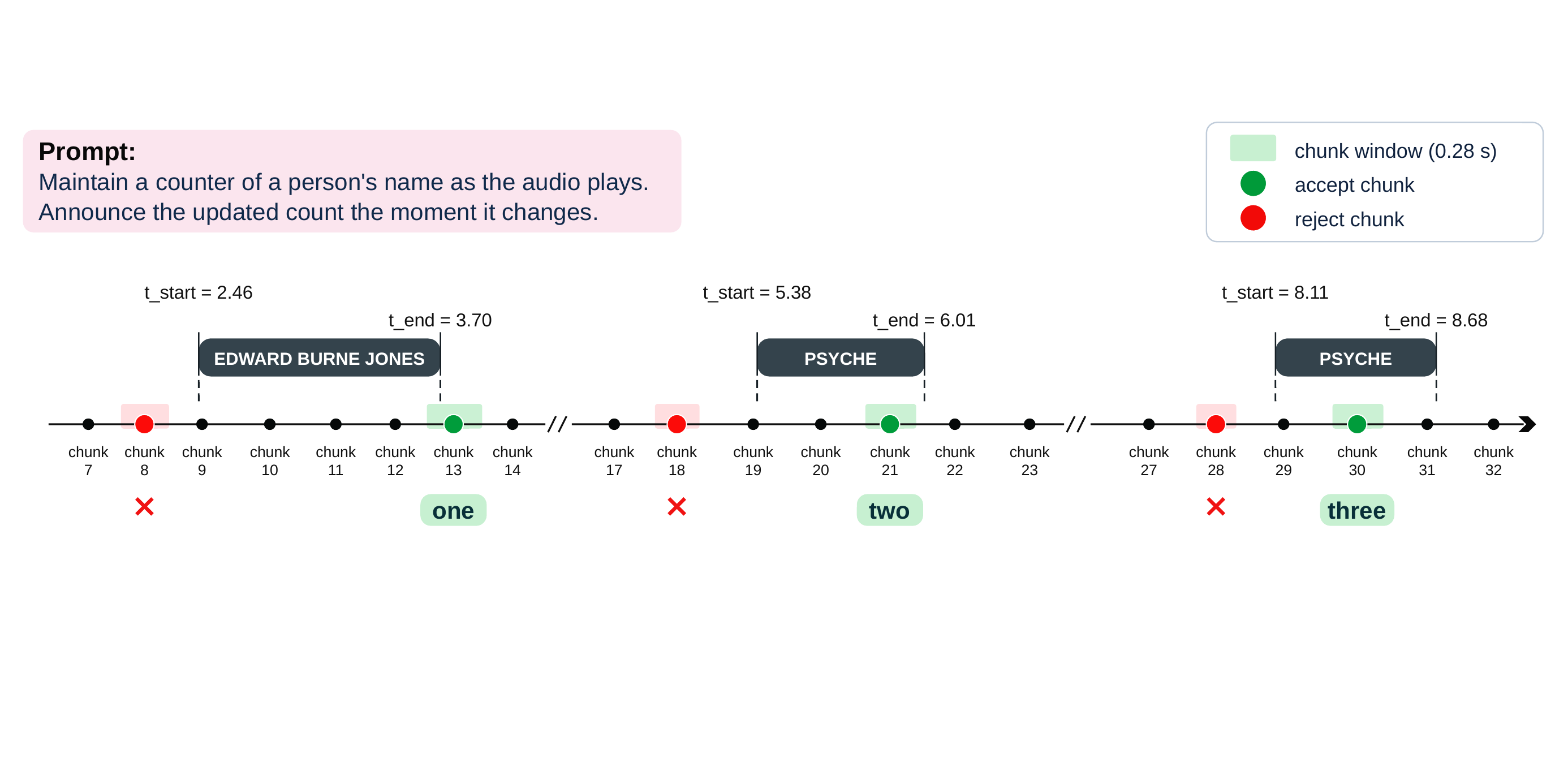}
\caption{Single-axis chunk timeline for person-name counting with a 0.28\,s chunk window. }
\label{fig:person-name-counter}
\end{figure*}

For lexical and semantic detection tasks, the model is expected to emit the target word when an event becomes observable; for counting tasks, it is expected to emit the updated cumulative count at each event (e.g.\ ``1'', ``2'', ``3''), and outputs
\linebreak[3]\texttt{<keep\_silence>}
otherwise. At each chunk boundary, the model implicitly selects one of two actions:
\textsc{SPEAK} or \textsc{SILENCE}. A \textsc{SPEAK} decision emits a meaningful content, followed by the end-of-turn token
\texttt{<|im\_end|>}, whereas a \textsc{SILENCE} decision emits the
special token \texttt{<keep\_silence>} followed by
\texttt{<|im\_end|>}. For an audio event whose end time is $e$, the chunk index is
\begin{equation}
k^{*}(e)=
\left\lceil\frac{e-\epsilon}{C}\right\rceil-1,
\qquad
0 \leq (k^{*}+1)C-e < C,
\label{eq:trigger}
\end{equation}
where $\epsilon \to 0^{+}$ denotes an infinitesimal positive time offset used to resolve exact-boundary cases: if $e=mC$, the event is assigned to the chunk whose right boundary is $mC$, rather than to the following chunk. This ensures that the supervision is causal and introduces less than one chunk of delay. At inference each emission is assigned to the same boundary $(k^{*}\!+\!1)C$, making the recorded onset the causal decision time.
Figure~\ref{fig:person-name-counter} illustrates the resulting single-axis chunk timeline for semantic counting of person names, where chunks overlapping the annotated name span are accepted and the rest rejected.

Loss is computed only over outputs corresponding to the streaming speech, with output positions corresponding to the system prompt masked. A positive sample contains one or more \textsc{SPEAK} decisions, while the remaining chunks correspond to \textsc{SILENCE} decisions, so directly averaging token-level cross-entropy strongly biases the model toward an always-silent solution. We therefore separate the loss into four groups---\textsc{SPEAK} content, \textsc{SPEAK} end token, \textsc{SILENCE} action, and \textsc{SILENCE} end token---and average the two components within each decision type, giving $\mathcal{L}_{s}=\tfrac{1}{2}(\bar{\ell}_{\mathrm{spk}}+\bar{\ell}_{\mathrm{spk,end}})$ and
$\mathcal{L}_{q}=\tfrac{1}{2}(\bar{\ell}_{\mathrm{sil}}+\bar{\ell}_{\mathrm{sil,end}})$,
where $\bar{\ell}$ is the mean token cross-entropy over a bucket
($\mathrm{spk}/\mathrm{sil}$: \texttt{<speak>}/ \texttt{<keep\_silence>}, $\mathrm{end}$: \texttt{<|im\_end|>})
and $w_s,w_q$ the balancing weights. We balance the two:
\begin{equation}
\mathcal{L} = \frac{w_s\mathcal{L}_s+w_q\mathcal{L}_q}{w_s+w_q},
\qquad w_s=w_q=0.5,
\label{eq:balanced_loss}
\end{equation}
\label{eq:balanced_loss}
so that SPEAK and SILENCE have equal aggregate importance regardless of frequency, while separating the \texttt{<|im\_end|>} loss keeps the sequence-ending signal from being diluted by content length.

We also utilize negative samples as described in Sec~\ref{sec:data} during training. It teaches the model to remain silent when the audio stream does not contain the queried target. 
We freeze the audio encoder and disable the speech-generation talker branch. LoRA adapters~\cite{DBLP:conf/iclr/HuSWALWWC22} are applied to the thinker attention and FFN projections (rank $r=16$, $\alpha=32$, dropout $0.05$), while modules remain fully trainable so that special tokens such as \texttt{<keep\_silence>} can be learned. We train with AdamW~\cite{DBLP:conf/iclr/LoshchilovH19} at $1.5\times10^{-4}$ with linear decay, batch size 96, bf16 precision, and an 8192-token context window.
\section{Experiments}
\label{sec:experiments}

\subsection{Setup}
\label{sec:setup}
We evaluate three groups of models on LibriSpeech-Real-QA. The first
comprises proprietary APIs for real-time speech interaction: Gemini
Live 3.1 Flash~\cite{google_gemini_live},
GPT-realtime-2.1-mini~\cite{openai_gpt_realtime},
qwen-audio-3.0-realtime-flash~\cite{qwen_audio_realtime}, and
qwen-audio-3.0-realtime-plus~\cite{qwen_audio_realtime}. The second
comprises pretrained Qwen2.5-Omni-7B~\cite{DBLP:journals/corr/abs-2503-20215}
and a version finetuned for offline inference, denoted \textsc{Offline}.
In the offline formulation, the model receives the full audio input and
is instructed to produce a timestamped event list in \texttt{JSON}
format. The third group, denoted \textsc{Streaming}, adapts
Qwen2.5-Omni-7B as described in
Sec.~\ref{sec:streaming_formulation}. We evaluate streaming models
with chunk durations of $400$, $280$, and $160$\,ms. We report results for the grace period $g=+1c$ and $g=+2c$ indicating the extension of the acceptance window with an extra one chunk and two chunks, respectively.

For the finetuned models, fuzzy and exact matching give nearly
identical results across the four tasks. The pretrained model shows a
larger discrepancy, as its responses more often include additional
contextual words. We therefore use fuzzy matching throughout for a
consistent comparison across model groups.
\subsection{Baseline Results}
\label{sec:results}

Tables~\ref{tab:lexical_detection}--\ref{tab:semantic_counting} report results on the four tasks for the \texttt{test-clean} and \texttt{test-other} splits. Existing real-time speech APIs can identify some queried events, particularly in lexical detection, but do not exhibit the event-triggered response behavior required by our tasks. Their mean onset delays are several seconds, consistent with treating the input as a conversational turn and responding near the end of the utterance rather than when the queried event occurs. These delays are not a controlled latency comparison across providers, as they also reflect differences in response policies and service conditions. Instead, the results demonstrate the value of LibriSpeech-Real-QA: by evaluating both \emph{what} a model emits and \emph{when} it emits it, the benchmark reveals a gap in fine-grained streaming speech interaction that content-only or turn-level evaluation would miss.

The pretrained Qwen2.5-Omni-7B has the lowest F1 on every task and
split, indicating that general speech capabilities alone do not suffice
for these event-triggered tasks. Finetuning for offline inference
substantially improves event matching: \textsc{Offline} achieves F1 of
$0.94/0.91$ on lexical detection, $0.98/0.96$ on lexical counting,
$0.78/0.66$ on semantic detection, and $0.91/0.89$ on semantic
counting (\texttt{test-clean}/\texttt{test-other}). 
However, we should note that the \textsc{Offline} variant treats the real-time speech understanding task as a post-audio target identification and localization problem.  
It thus can only serve as an reference and should not be compared directly with those of \textsc{Streaming}. 

\begin{table}[htb]
\centering
\caption{Lexical detection on LibriSpeech-Real-QA. $^{\dagger}$ indicates models that operate with the post-audio inference mode.}
\label{tab:lexical_detection}
\setlength{\tabcolsep}{2.5pt}
\resizebox{\columnwidth}{!}{%
\begin{tabular}{llccc c}
\toprule
Model & Win. & F1 & Recall & Precision & $\overline{d}$ \\
\midrule
Gemini Live 3.1 Flash & -- & 0.83~/~0.87 & 0.98~/~0.96 & 0.72~/~0.79 & 5.00~/~4.35 \\
GPT-realtime-2.1-mini & -- & 0.94~/~0.83 & 0.98~/~0.88 & 0.91~/~0.79 & 5.43~/~6.52 \\
Qwen-Audio Flash      & -- & 0.79~/~0.81 & 0.78~/~0.81 & 0.80~/~0.81 & 4.85~/~5.22 \\
Qwen-Audio Plus       & -- & 0.78~/~0.79 & 0.78~/~0.79 & 0.79~/~0.78 & 4.84~/~5.47 \\
\midrule
Qwen2.5-Omni-7B$^{\dagger}$       & -- & 0.50~/~0.51 & 0.50~/~0.50 & 0.51~/~0.52 & 1.457~/~1.328 \\
\textsc{Offline}$^{\dagger}$      & -- & 0.94~/~0.91 & 0.93~/~0.91 & 0.94~/~0.92 & 0.010~/~0.003 \\
\midrule
\multirow{2}{*}{\textsc{Streaming} (400\,ms)}
& +1c
& \textbf{0.988~/~0.968}
& 0.990~/~0.970
& \textbf{0.986~/~0.965}
& 0.660~/~0.623 \\

& +2c
& \textbf{0.993~/~0.978}
& 0.994~/~0.981
& \textbf{0.991~/~0.976}
& 0.661~/~0.625 \\

\multirow{2}{*}{\textsc{Streaming} (280\,ms)}
& +1c
& 0.984~/~0.967
& \textbf{0.991~/~0.972}
& 0.978~/~0.963
& 0.585~/~0.542 \\

& +2c
& 0.990~/~0.975
& \textbf{0.996~/~0.980}
& 0.983~/~0.971
& 0.586~/~0.543 \\

\multirow{2}{*}{\textsc{Streaming} (160\,ms)}
& +1c
& 0.982~/~0.959
& 0.987~/~0.962
& 0.976~/~0.957
& \textbf{0.513~/~0.472} \\

& +2c
& 0.985~/~0.966
& 0.991~/~0.968
& 0.980~/~0.963
& \textbf{0.513~/~0.473} \\
\bottomrule
\end{tabular}%
}
\end{table}

\begin{table}[htb]
\centering
\caption{Lexical counting on LibriSpeech-Real-QA}
\label{tab:lexical_counting}
\setlength{\tabcolsep}{3.5pt}
\resizebox{\columnwidth}{!}{%
\begin{tabular}{llccc c}
\toprule
Model & Win. & F1 & Recall & Precision & $\overline{d}$ \\
\midrule
Gemini Live 3.1 Flash & -- & 0.61~/~0.68 & 0.68~/~0.68 & 0.56~/~0.69 & 8.94~/~7.15 \\
GPT-realtime-2.1-mini & -- & 0.58~/~0.46 & 0.45~/~0.36 & 0.85~/~0.61 & 10.07~/~10.03 \\
Qwen-Audio Flash      & -- & 0.44~/~0.42 & 0.32~/~0.34 & 0.67~/~0.54 & 4.03~/~6.52 \\
Qwen-Audio Plus       & -- & 0.44~/~0.45 & 0.32~/~0.35 & 0.68~/~0.64 & 4.07~/~7.03 \\
\midrule
Qwen2.5-Omni-7B$^{\dagger}$ & -- & 0.07~/~0.08 & 0.10~/~0.13 & 0.05~/~0.06 & 2.419~/~2.990 \\
\textsc{Offline}$^{\dagger}$
& -- & \textbf{0.98~/~0.96}
& \textbf{0.97~/~0.97}
& \textbf{0.99~/~0.96}
& 0.062~/~0.023 \\
\midrule
\multirow{2}{*}{\textsc{Streaming} (400\,ms)}
& +1c
& 0.952~/~0.923
& 0.960~/~0.935
& 0.944~/~0.912
& 0.585~/~0.531 \\
& +2c
& 0.954~/~0.932
& 0.962~/~0.944
& 0.946~/~0.920
& 0.586~/~0.533 \\

\multirow{2}{*}{\textsc{Streaming} (280\,ms)}
& +1c
& \textbf{0.962~/~0.944}
& \textbf{0.970~/~0.955}
& \textbf{0.954~/~0.933}
& 0.514~/~0.465 \\
& +2c
& \textbf{0.964~/~0.945}
& \textbf{0.972~/~0.957}
& \textbf{0.956~/~0.935}
& 0.515~/~0.465 \\

\multirow{2}{*}{\textsc{Streaming} (160\,ms)}
& +1c
& 0.926~/~0.882
& 0.940~/~0.900
& 0.914~/~0.864
& \textbf{0.458~/~0.400} \\
& +2c
& 0.928~/~0.884
& 0.942~/~0.902
& 0.916~/~0.866
& \textbf{0.458~/~0.400} \\
\bottomrule
\end{tabular}%
}
\end{table}
\begin{table}[htb]
\centering
\caption{Semantic detection on LibriSpeech-Real-QA}
\label{tab:semantic_detection}
\setlength{\tabcolsep}{3.5pt}
\resizebox{\columnwidth}{!}{%
\begin{tabular}{llccc c}
\toprule
Model & Win. & F1 & Recall & Precision & $\overline{d}$ \\
\midrule
Gemini Live 3.1 Flash & -- & 0.65~/~0.61 & 0.70~/~0.65 & 0.61~/~0.57 & 5.89~/~5.09 \\
GPT-realtime-2.1-mini & -- & 0.73~/~0.44 & 0.72~/~0.45 & 0.73~/~0.44 & 6.64~/~6.66 \\
Qwen-Audio Flash & -- & 0.52~/~0.45 & 0.49~/~0.44 & 0.56~/~0.46 & 3.25~/~6.13 \\
Qwen-Audio Plus & -- & 0.54~/~0.47 & 0.51~/~0.46 & 0.57~/~0.50 & 3.23~/~6.32 \\
\midrule
Qwen2.5-Omni-7B$^{\dagger}$
& -- & 0.25~/~0.24 & 0.26~/~0.23 & 0.25~/~0.25 & 1.758~/~1.716 \\
\textsc{Offline}$^{\dagger}$
& -- & 0.78~/~0.66 & 0.75~/~0.64 & 0.81~/~0.69 & 0.021~/~0.013 \\
\midrule
\multirow{2}{*}{\textsc{Streaming} (400\,ms)}
& +1c
& \textbf{0.824~/~0.694}
& \textbf{0.825~/~0.686}
& \textbf{0.823~/~0.701}
& 0.637~/~0.614 \\
& +2c
& \textbf{0.828~/~0.707}
& \textbf{0.829~/~0.700}
& \textbf{0.827~/~0.715}
& 0.639~/~0.619 \\

\multirow{2}{*}{\textsc{Streaming} (280\,ms)}
& +1c & 0.776~/~0.661 & 0.799~/~0.666 & 0.754~/~0.656 & 0.558~/~0.534 \\
& +2c & 0.781~/~0.670 & 0.804~/~0.675 & 0.759~/~0.665 & 0.561~/~0.537 \\

\multirow{2}{*}{\textsc{Streaming} (160\,ms)}
& +1c
& 0.721~/~0.583
& 0.754~/~0.598
& 0.692~/~0.568
& \textbf{0.476~/~0.450} \\
& +2c
& 0.729~/~0.585
& 0.761~/~0.601
& 0.698~/~0.570
& \textbf{0.477~/~0.451} \\
\bottomrule
\end{tabular}%
}
\end{table}

\begin{table}[htb]
\centering
\caption{Semantic counting on LibriSpeech-Real-QA}
\label{tab:semantic_counting}
\setlength{\tabcolsep}{3.5pt}
\resizebox{\columnwidth}{!}{%
\begin{tabular}{llccc c}
\toprule
Model & Win. & F1 & Recall & Precision & $\overline{d}$ \\
\midrule
Gemini Live 3.1 Flash & -- & 0.49~/~0.54 & 0.59~/~0.54 & 0.42~/~0.54 & 8.86~/~8.04 \\
GPT-realtime-2.1-mini & -- & 0.48~/~0.45 & 0.37~/~0.35 & 0.69~/~0.65 & 10.21~/~11.27 \\
Qwen-Audio Flash      & -- & 0.40~/~0.43 & 0.30~/~0.35 & 0.61~/~0.55 & 4.13~/~7.20 \\
Qwen-Audio Plus       & -- & 0.43~/~0.48 & 0.32~/~0.37 & 0.68~/~0.69 & 4.11~/~7.60 \\
\midrule
Qwen2.5-Omni-7B$^{\dagger}$ & -- & 0.04~/~0.02 & 0.04~/~0.01 & 0.05~/~0.03 & 3.074~/~1.748 \\
\textsc{Offline}$^{\dagger}$
& -- & 0.91~/~0.89 & 0.89~/~0.86 & 0.92~/~0.92 & 0.220~/~0.128 \\
\midrule
\multirow{2}{*}{\textsc{Streaming} (400\,ms)}
& +1c
& \textbf{0.848~/~0.721}
& \textbf{0.863~/~0.761}
& \textbf{0.833~/~0.684}
& 0.623~/~0.603 \\
& +2c
& \textbf{0.850~/~0.733}
& \textbf{0.865~/~0.774}
& \textbf{0.835~/~0.696}
& 0.624~/~0.609 \\

\multirow{2}{*}{\textsc{Streaming} (280\,ms)}
& +1c & 0.809~/~0.709 & 0.827~/~0.755 & 0.791~/~0.669 & 0.568~/~0.531 \\
& +2c & 0.814~/~0.716 & 0.833~/~0.762 & 0.797~/~0.676 & 0.569~/~0.533 \\

\multirow{2}{*}{\textsc{Streaming} (160\,ms)}
& +1c
& 0.773~/~0.618
& 0.803~/~0.670
& 0.745~/~0.573
& \textbf{0.485~/~0.464} \\
& +2c
& 0.778~/~0.628
& 0.808~/~0.681
& 0.750~/~0.583
& \textbf{0.486~/~0.466} \\
\bottomrule
\end{tabular}%
}
\end{table}
\begin{table}[htb]
\centering
\caption{Positive:negative sampling-ratio ablation of the 280\,ms streaming
model on the \texttt{test-clean} split of LibriSpeech-Real-QA
(compute-matched, 1 epoch, \emph{+1c} window).}
\label{tab:ablation_posneg}
\small
\setlength{\tabcolsep}{4pt}
\resizebox{0.6\columnwidth}{!}{%
\begin{tabular}{@{}llccc@{}}
\toprule
pos:neg & Task & F1 & Recall & Precision \\
\midrule
\multirow{4}{*}{1.0\,{:}\,0.0}
& Lexical det.  & 0.952 & 0.960 & 0.943 \\
& Semantic det. & 0.715 & 0.751 & 0.683 \\
& Lexical cnt.  & 0.926 & 0.944 & 0.909 \\
& Semantic cnt. & 0.724 & 0.780 & 0.676 \\
\midrule
\multirow{4}{*}{0.9\,{:}\,0.1}
& Lexical det.
& \textbf{0.972} & \textbf{0.979} & \textbf{0.966} \\
& Semantic det.
& \textbf{0.750} & \textbf{0.776} & \textbf{0.727} \\
& Lexical cnt.  & 0.928 & 0.946 & 0.911 \\
& Semantic cnt. & 0.793 & 0.833 & 0.757 \\
\midrule
\multirow{4}{*}{0.8\,{:}\,0.2}
& Lexical det.  & 0.970 & 0.978 & 0.962 \\
& Semantic det. & 0.736 & 0.762 & 0.713 \\
& Lexical cnt.
& \textbf{0.935} & \textbf{0.950} & \textbf{0.920} \\
& Semantic cnt.
& \textbf{0.805} & \textbf{0.844} & \textbf{0.769} \\
\bottomrule
\end{tabular}
}
\end{table}


\subsection{Streaming Finetuning Results}
\label{sec:analysis}
\noindent\textbf{Overall comparison.}
Streaming finetuning enables Qwen2.5-Omni-7B to respond during the
speech stream and improves F1 over the pretrained model on every task
and split. It also outperforms the evaluated real-time APIs in F1
across all four tasks. Under the \emph{+1c} window, the $400$\,ms
streaming model achieves $0.988/0.968$ F1 on lexical detection and
$0.824/0.694$ on semantic detection, exceeding the corresponding
\textsc{Offline} results while making causal decisions. In contrast,
\textsc{Offline} remains stronger on both counting tasks. These
results establish a strong causal baseline and show why the
benchmark must evaluate response timing alongside answer content.

\noindent\textbf{Task-wise comparison.}
For our finetuned \textsc{Streaming} model, the semantic tasks generally yield lower F1 than the corresponding lexical tasks, indicating that they are more challenging in this setting. This is expected because semantic tasks require the model to relate the speech signal to the meaning of the queried class, whereas lexical tasks require it to recognize a specified word.
Comparing detection and counting reveals different patterns across the two task families. For lexical tasks, counting is more challenging than detection; for the finetuned streaming models, semantic detection has lower F1 than semantic counting.
For lexical tasks, detecting a repeated target word is sufficient for detection, but counting also requires the model to maintain and update a cumulative count after each occurrence. For semantic tasks, both tasks require recognizing class membership, but detection must additionally identify and output the particular word heard at each event; counting only needs to output the updated count.

\noindent\textbf{Effect of chunk size.}
Chunk duration exposes a task-dependent accuracy--delay trade-off.
Under the \emph{+2c} window, the $400$\,ms model achieves the
highest streaming F1 on lexical detection ($0.993/0.978$), semantic
detection ($0.828/0.707$), and semantic counting ($0.850/0.733$).
The $280$\,ms model performs best on lexical counting
($0.964/0.945$), while the $160$\,ms model has the lowest mean onset
delay on every task and split. For lexical detection under
\emph{+1c}, reducing chunk duration from $400$ to $280$ to
$160$\,ms lowers mean onset delay on \texttt{test-clean} from
$0.660$ to $0.585$ to $0.513$\,s, while F1 changes from $0.988$
to $0.984$ to $0.982$. F1 falls more sharply with shorter chunks
on the semantic tasks. Thus, the benchmark reveals not just whether
a model can respond causally, but how promptly it can do so at a
given level of task performance.

\noindent\textbf{Different evaluation window sizes.}
Extending the onset window from \emph{+1c} to \emph{+2c} increases
streaming F1 by at most $0.013$ across the reported tasks, chunk
durations, and splits. The wider window credits some later emissions
but does not substantially change the pattern of results.

\subsection{Ablation Study on Negative Sampling}
Table~\ref{tab:ablation_posneg} examines the positive-to-negative
sampling ratio for the streaming model. A $0.9{:}0.1$ ratio yields the
highest F1 on lexical and semantic detection ($0.972$ and $0.750$),
whereas $0.8{:}0.2$ performs best on lexical and semantic counting
($0.935$ and $0.805$). Removing negative samples produces the
lowest F1 on all four tasks.
\section{Conclusion}
\label{sec:conclusion}

We introduced LibriSpeech-Real-QA, a benchmark for event-triggered
streaming speech interaction. Its temporal evaluation protocol
assesses both the content and timing of model responses, revealing
that existing real-time speech APIs often respond near the end of an
utterance rather than when a queried event occurs. We also showed
that streaming finetuning can adapt a pretrained SLM to make causal,
event-level responses, establishing baselines across chunk durations
and exposing task-dependent accuracy--delay trade-offs. The benchmark
provides a basis for developing speech models that answer while
speech is still unfolding.

\vfill\pagebreak

\bibliographystyle{IEEEbib}
\bibliography{strings,refs}

\end{document}